\documentclass[letterpaper, 10 pt, conference]{ieeeconf}  % Comment this line out if you need a4paper

\IEEEoverridecommandlockouts                              % This command is only needed if 
\usepackage{amsmath}                                                          
\usepackage{amssymb} % 支持 \checkmark
\usepackage{xcolor}  % 彩色
\usepackage{pifont}  % 支持 \ding{55} 叉号
\usepackage{multirow}
 \usepackage{graphicx}
\usepackage{tikz}
\usepackage{makecell} % 导言区加

\title{\LARGE \bf
PERAL: Perception-Aware Motion Control for Passive LiDAR Excitation in Spherical Robots
}

\author{Shenghai Yuan$^{*}$, Jason Wai Hao Yee$^{*}$, Weixiang Guo, Zhongyuan Liu, Thien-Minh Nguyen, and Lihua Xie%
\thanks{$^{*}$Equal Contribution.}%
\thanks{All authors are with the Centre for Advanced Robotics Technology Innovation (CARTIN), School of Electrical and Electronic Engineering, Nanyang Technological University, 50 Nanyang Avenue, Singapore 639798.}%
\thanks{Emails: \{shyuan, elhxie\}@ntu.edu.sg}%
}

\begin{document}

\maketitle
\thispagestyle{empty}
\pagestyle{empty}

%%%%%%%%%%%%%%%%%%%%%%%%%%%%%%%%%%%%%%%%%%%%%%%%%%%%%%%%%%%%%%%%%%%%%%%%%%%%%%%%
\begin{abstract}
Autonomous mobile robots increasingly rely on LiDAR–IMU odometry for navigation and mapping, yet horizontally mounted LiDARs such as the MID360 capture few near-ground returns, limiting terrain awareness and degrading performance in feature-scarce environments. Prior solutions—static tilt, active rotation, or high-density sensors—either sacrifice horizontal perception or incur added actuators, cost, and power.
We introduce PERAL, a perception-aware motion control framework for spherical robots that achieves passive LiDAR excitation without dedicated hardware. By modeling the coupling between internal differential-drive actuation and sensor attitude, PERAL superimposes bounded, non-periodic oscillations onto nominal goal- or trajectory-tracking commands, enriching vertical scan diversity while preserving navigation accuracy.
Implemented on a compact spherical robot, PERAL is validated across laboratory, corridor, and tactical environments. Experiments demonstrate up to 96\% map completeness, a 27\% reduction in trajectory tracking error, and robust near-ground human detection, all at lower weight, power, and cost compared with static tilt, active rotation, and fixed horizontal baselines. The design and code will be open-sourced upon acceptance.
\end{abstract}

%%%%%%%%%%%%%%%%%%%%%%%%%%%%%%%%%%%%%%%%%%%%%%%%%%%%%%%%%%%%%%%%%%%%%%%%%%%%%%%%
\section{INTRODUCTION}
LiDAR-based perception \cite{hou2025enhancing} has become a cornerstone technology in mobile robotics \cite{luo2023bevplace,luo2025bevplace++}, enabling accurate mapping \cite{zhong20243d}, localization \cite{yuan2021survey}, and navigation \cite{yin2024survey} in complex environments. Recent advances in lightweight 3D LiDARs \cite{yang2025fast}, such as the MID360 \cite{Zhu2022SwarmLIO}, have expanded their adoption in both autonomous vehicles and indoor service robots. However, a persistent challenge remains: ground coverage. Many LiDAR configurations—particularly those with horizontal mounting—fail to capture sufficient near-field ground points, leading to incomplete maps, degraded terrain understanding, and reduced SLAM accuracy \cite{lee2024switch} in flat or feature-sparse environments.

\textbf{Existing} solutions attempt to address this limitation in several ways. Actuated excitation mechanisms, as seen in works such as PULSAR \cite{Chen2023PULSAR}, UA-MPC \cite{Li2025UAMPC}, rotate the LiDAR periodically to expand the vertical field-of-view and improve ground visibility. While effective, these approaches introduce additional actuators, control complexity, and power consumption. Alternatively, static tilted mounting \cite{zhang2025yuto} offers a simple, low-cost improvement in ground coverage, but sacrifices long-range horizontal scanning uniformity and may degrade loop closure performance. The tilted ways are more likely to deteriorate in a long corridor area \cite{pfreundschuh2024coin}. High-density LiDARs like the Ouster series inherently capture more ground points but remain cost- and power-prohibitive for compact platforms.

The core \textbf{challenge} lies in increasing ground coverage without compromising horizontal perception, adding mechanical complexity, or incurring significant energy cost. Achieving this requires rethinking how sensor motion can be enhanced without dedicated actuators, and how perception can be improved as a byproduct of natural platform dynamics.
\begin{figure}[t]
    \centering
    \includegraphics[width=0.95\linewidth]{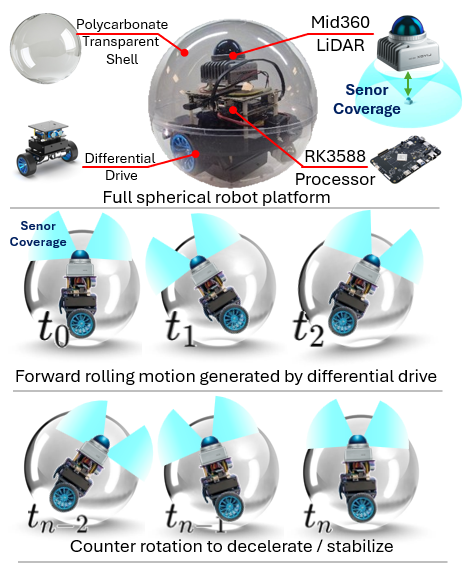}
    \caption{Hardware and perception architecture of the proposed spherical robot (top) and its rolling motion over time (bottom). The polycarbonate shell provides protection, while the internal differential drive enables rolling locomotion. A Mid360 LiDAR mounted on the top provides 360° 3D perception, fused by the onboard RK3588 processor for mapping and situational awareness. The time sequence illustrates rolling acceleration (“Speed Up”) and controlled deceleration (“Braking”). The pitch-up and down motion enables wider LiDAR coverage without an additional motor.}
    \label{fig:motivation}
\end{figure}
In this work, we present a novel approach: passive LiDAR excitation through the intrinsic motion of spherical mobile robots. In our design, the drive mechanism—wheels operating inside the spherical shell—naturally induces small, non-periodic attitude perturbations of the LiDAR during locomotion. These passive orientation changes broaden the vertical scanning distribution, significantly increasing near-field ground visibility without additional actuators, control modules, or power draw. We validate our method through quantitative experiments comparing ground point coverage, SLAM accuracy, and mapping completeness against conventional horizontal mounting, static tilt, and active rotation baselines.

\noindent\textbf{The main contributions of this work are:}
\begin{itemize}
    \item We develop a perception-oriented motion model for spherical robots that captures the coupling between internal actuation and LiDAR orientation changes, enabling controlled attitude oscillations for simultaneous trajectory tracking and ground coverage enhancement.
    
    \item We propose a control strategy that superimposes bounded, model-based oscillatory motions on nominal trajectory commands, improving vertical scanning diversity and LiDAR–IMU observability without sacrificing navigation accuracy or requiring additional actuators.
    
    \item We implement and validate the proposed approach on a real spherical robot platform, demonstrating significant improvements in ground point coverage, mapping completeness, and SLAM accuracy compared to horizontal mounting, static tilt, and active rotation baselines.
\end{itemize}

\noindent \textbf{Remark:} Here, \emph{observability} denotes the practical extent to which robot motion excites the LiDAR field-of-view so that critical structures (e.g., near-ground points, vertical planes) are captured. This differs from classical EKF observability, as we do not alter the Jacobian rank but enhance point coverage and spatial distribution, thereby improving map completeness and trajectory accuracy.

\section{Related Work}

\subsection{LiDAR-Based SLAM and Perception Limitations}

LiDAR-based simultaneous localization \cite{xu2025enhancing} and mapping (SLAM) \cite{zhang2014loam,chen2021range,chen2019suma++, chen2021ndt,wang2021f,li2025graph,chen2025egs} has advanced rapidly with the emergence of tightly coupled LiDAR–inertial odometry (LIO) frameworks \cite{koide2022globally,koide2024glim,Lou2025QLIO}. Representative works include \emph{FAST-LIO} and its successors~\cite{Xu2020FASTLIO,Xu2021FASTLIO2}, which exploit iterated Kalman filtering and direct raw point registration for efficient and robust state estimation. \emph{Swarm-LIO}~\cite{Zhu2022SwarmLIO} extends this paradigm to multi-UAV scenarios, enabling decentralized odometry under bandwidth and scalability constraints. More recently, \emph{IG-LIO}~\cite{IGLIO2023} extends this line of work to large-scale and long-duration scenarios, leveraging an incremental GICP formulation for tightly coupled LiDAR–IMU odometry.
 Beyond pure LiDAR–IMU systems, multi-modal approaches such as \emph{R3LIVE}~\cite{Lin2022R3LIVE} introduce visual and color cues for improved perceptual richness, while \emph{BEV-LIO(LC)}~\cite{BEVLIO2024} leverages bird’s-eye-view image priors to facilitate loop closure. Similarly, \emph{GLIM}~\cite{Koide2022GLIM} employs GPU-accelerated generalized ICP factors for scalable scan matching, and \emph{LiTAMIN2}~\cite{Yokozuka2021LiTAMIN2} achieves lightweight mapping through KL-divergence based geometric approximations. Finally, \emph{Adaptive-LIO}~\cite{AdaptiveLIO2024} demonstrates that environment-aware adaptation can further stabilize odometry across diverse scenes.

Despite these advances, a fundamental limitation persists: horizontally mounted LiDARs often provide insufficient vertical coverage on compact platforms. In feature-sparse settings—such as corridors, open fields, or smoke-filled spaces—this results in degraded observability, incomplete maps, and fragile loop closures. While recent algorithms improve efficiency and robustness, they cannot fully compensate for sensing blind spots, motivating research into motion-induced strategies that enrich LiDAR coverage.

\subsection{Active Excitation Mechanisms for Observability}

To overcome the sensing blind spots of horizontally mounted LiDARs, a number of works have pursued \emph{active excitation}, where additional motions or actuators are introduced to enrich scan diversity and improve state observability. Representative examples include \emph{Point-LIO}~\cite{He2022PointLIO}, which achieves high-bandwidth LiDAR–IMU odometry by leveraging rapid motion and low-latency sensor fusion, thereby maintaining performance in aggressive trajectories. Similarly, \emph{PULSAR}~\cite{Chen2023PULSAR} presents a self-rotating, single-actuated UAV that continuously spins to extend its field of view for navigation and obstacle detection. 

Another line of work incorporates active control strategies. For instance, \emph{UA-MPC}~\cite{Li2025UAMPC} formulates LiDAR odometry within an uncertainty-aware model predictive control framework, where actuator-driven oscillations explicitly increase vertical coverage. In parallel, calibration-oriented efforts such as \emph{LiMo-Calib}~\cite{Li2023LiMoCalib} explore motorized LiDAR setups on quadruped robots, enabling panoramic 3D sensing through deliberate rotational excitation.

While effective, these active excitation methods inevitably introduce drawbacks: they require additional actuators \cite{li2025aeos} or control complexity, increase power consumption, and often demand precise calibration or synchronization. Such trade-offs limit their practicality for lightweight and energy-constrained robotic platforms.

Koide et al.~\cite{Koide2022gc} developed a system for globally consistent LiDAR–inertial mapping using GPU-accelerated registration. In later extensions, they demonstrated a passive spherical robot prototype, indicating that platform dynamics alone can be exploited to enhance perception—an idea that directly inspired our actuator-free approach.

% \subsection{Ground Coverage Enhancement Strategies}
% Several strategies have been developed to address the lack of near-field ground coverage in horizontally mounted LiDAR systems. \textit{Static tilt mounting}~\cite{park2021tiltedLiDAR} offers a simple, low-cost solution by angling the LiDAR downwards to capture more ground points. While effective in certain scenarios, this approach reduces uniform horizontal coverage and can lead to uneven point density across different ranges, potentially harming long-range mapping quality. 
% \textit{Active excitation mechanisms} provide a more dynamic solution. Approaches such as UA-MPC~\cite{li2022uamapc} mount the LiDAR on a motorized gimbal to oscillate its orientation periodically, enriching vertical scanning diversity and improving SLAM robustness. Similar strategies have been used in autonomous driving to detect low obstacles~\cite{kim2020rotatingLiDAR}. These methods, however, increase mechanical complexity, require precise synchronization between actuator motion and sensor sampling, and add both weight and energy consumption.
% Finally, \textit{high-channel LiDARs} like the Ouster OS series capture more vertical beams by design, improving ground coverage without mechanical excitation. Yet, their cost, size, and power requirements limit adoption in small-scale mobile platforms.
\begin{figure*}[t]
    \centering
    \includegraphics[width=1\linewidth]{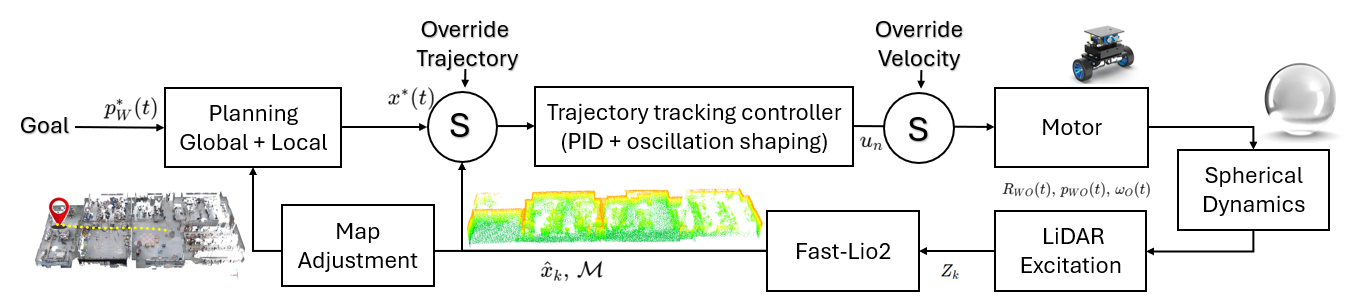}
    \caption{System architecture of the spherical robot with multi-mode control inputs. The framework supports goal-driven planning, direct trajectory commands, and manual velocity overrides, all unified through a trajectory tracking controller with oscillation shaping. The resulting motion naturally excites the LiDAR for enhanced perception without additional actuation.}
    \label{fig:flowchart}
\end{figure*}

\subsection{Persistent Excitation and Observability}

The concept of \emph{persistent excitation} (PE) is fundamental for ensuring observability and stability in estimation and control. Recent theoretical studies formalize explicit PE requirements for observer stability~\cite{Bouazza2024} and analyze global observability conditions in nonholonomic robotic systems~\cite{Palopoli2020,huang2025geometric,yang2025integrated,setayeshi2025terminal,zhao2025adaptive,wang2023integrated,wen2020hilps}. 

In multi-robot and swarm scenarios, it has been shown that persistently exciting relative motions enable accurate adaptive localization and stable formation control~\cite{Nguyen2020}. Related works on target encirclement further exploit motion diversity to improve robustness against noise and environmental complexity~\cite{Liu2020Automatica,Liu2021TCST,Liu2021TIE}. These results highlight that perception and control are tightly coupled, and that observability can be actively enhanced by introducing motion richness. 

However, enforcing PE often requires deliberate trajectory design or additional actuation, which may be costly for small, resource-constrained robots. This motivates the exploration of \emph{passive motion coupling}, where natural dynamics—such as rolling, braking, and turning—provide sufficient excitation to diversify LiDAR scans and enhance near-ground observability without dedicated mechanisms.

\section{Proposed Solutions}

\subsection{System Overview}
As shown in Fig. \ref{fig:flowchart}, the spherical robot navigates either toward assigned goals via global–local planning or along predefined trajectories through its internal differential-drive unit. Natural acceleration, braking, and turning induce small pitch–roll oscillations of the outer shell, tilting the horizontally mounted LiDAR and passively diversifying scan directions without extra actuation. LiDAR–IMU fusion is performed using the discrete-time FAST-LIO pipeline, with LiDAR frames processed at their acquisition rate and IMU preintegration in between. Hardware, calibration, and baseline settings are detailed in Sec.~\ref{sec:exp_setup}.

\subsection{Problem Formulation}

\noindent \textbf{Frames and Extrinsics:}
Let $\{\mathcal{W}\}$ be world, $\{\mathcal{O}\}$ the outer-shell frame at the sphere center,
$\{\mathcal{I}\}$ the internal drive frame, and $\{\mathcal{L}\}$ the LiDAR frame.
Shell pose:
\begin{equation}
    \mathbf{T}_{\mathcal{W}\mathcal{O}}(t)=
    \begin{bmatrix}
        \mathbf{R}_{\mathcal{W}\mathcal{O}}(t) & \mathbf{p}_{\mathcal{W}\mathcal{O}}(t)\\
        \mathbf{0}^\top & 1
    \end{bmatrix}.
\end{equation}
Rigid extrinsics:
\begin{equation}
    \mathbf{T}_{\mathcal{O}\mathcal{L}}=
    \begin{bmatrix}
        \mathbf{R}_{\mathcal{O}\mathcal{L}} & \mathbf{p}_{\mathcal{O}\mathcal{L}}\\
        \mathbf{0}^\top & 1
    \end{bmatrix},
    \mathbf{T}_{\mathcal{O}\mathcal{I}}(t)=
    \begin{bmatrix}
        \mathbf{R}_{\mathcal{O}\mathcal{I}}(t) & \mathbf{p}_{\mathcal{O}\mathcal{I}}(t)\\
        \mathbf{0}^\top & 1
    \end{bmatrix}.
\end{equation}

\noindent \textbf{State and IMU-Driven Dynamics:}
Let $\mathbf{x}(t) = [\mathbf{p}_{\mathcal{W}\mathcal{O}}(t),\,\mathbf{v}(t),\,\mathbf{R}_{\mathcal{W}\mathcal{O}}(t),\,\mathbf{b}_a(t),\,\mathbf{b}_\omega(t)]$ denote the platform state, where $\mathbf{p}_{\mathcal{W}\mathcal{O}}$ is the position of the sphere's body frame $\mathcal{O}$ in the world frame $\mathcal{W}$, $\mathbf{v}$ is the linear velocity, $\mathbf{R}_{\mathcal{W}\mathcal{O}}\in SO(3)$ is the orientation, and $\mathbf{b}_a$, $\mathbf{b}_\omega$ are IMU biases. 

The internal differential-drive kinematics are driven by the control input $\mathbf{u}(t) = [u_L(t), u_R(t)]^\top$:
\begin{equation}
    \mathbf{v}_{\mathcal{I}}(t) =
    \begin{bmatrix}
        \frac{r}{2}\big(u_L(t)+u_R(t)\big)\\
        0\\
        0
    \end{bmatrix},
\end{equation}
\begin{equation}
    \boldsymbol{\omega}_{\mathcal{I}}(t) =
    \begin{bmatrix}
        0\\
        0\\
        \frac{r}{d}\big(u_R(t)-u_L(t)\big)
    \end{bmatrix}.
\end{equation}

where $r$ is the wheel radius and $d$ is the track width of the internal drive unit.

Although the above dynamics are expressed in continuous time for clarity, in practice the FAST-LIO backend operates in discrete time: IMU measurements are preintegrated between LiDAR frames, and scan-to-map registration is performed at frame timestamps to estimate $\mathbf{x}(t_k)$ for each LiDAR frame time $t_k$.

\noindent \textbf{Internal Differential-Drive Kinematics:}
The internal actuation unit is a differential-drive mechanism mounted inside the spherical shell.  
Let $u_L(t)$ and $u_R(t)$ denote the left and right wheel angular velocities, $r$ the wheel radius, and $d$ the track width.  
The linear and angular velocities of the drive unit in its own frame $\{\mathcal{I}\}$ are:
\begin{align}
    \mathbf{v}_{\mathcal{I}}(t) &=
    \begin{bmatrix}
        \tfrac{r}{2}\big(u_L(t)+u_R(t)\big)\\[2pt]
        0\\[2pt]
        0
    \end{bmatrix}, \\
    \boldsymbol{\omega}_{\mathcal{I}}(t) &=
    \begin{bmatrix}
        0\\[2pt]
        0\\[2pt]
        \tfrac{r}{d}\big(u_R(t)-u_L(t)\big)
    \end{bmatrix}.
\end{align}

\noindent \textbf{Drive--Shell Coupling:}  
When the internal drive rolls without slip along the inner surface of the spherical shell (radius $R_s$),  
its motion induces an opposite translational velocity on the shell in the outer frame $\{\mathcal{O}\}$:
\begin{equation}
    \mathbf{v}_{\mathcal{O}}(t) = -\,\mathbf{R}_{\mathcal{O}\mathcal{I}}(t)\,\mathbf{v}_{\mathcal{I}}(t).
\end{equation}
The corresponding shell angular velocity is determined by the rolling constraint:
\begin{equation}
    \boldsymbol{\omega}_{\mathcal{O}}(t) = \frac{1}{R_s}\,\big(\mathbf{e}_z \times \mathbf{v}_{\mathcal{O}}(t)\big).
\end{equation}
Consequently, the shell pose evolves in the world frame $\{\mathcal{W}\}$ according to:
\begin{equation}
    \dot{\mathbf{p}}_{\mathcal{W}\mathcal{O}}(t) = \mathbf{v}_{\mathcal{O}}^{\mathcal{W}}(t),\quad
    \dot{\mathbf{R}}_{\mathcal{W}\mathcal{O}}(t) = \mathbf{R}_{\mathcal{W}\mathcal{O}}(t)\,\lfloor \boldsymbol{\omega}_{\mathcal{O}}(t)\rfloor.
\end{equation}

\noindent \textbf{LiDAR Pose and Measurement Model:}  
This shell motion directly affects the LiDAR extrinsics.  
With fixed mount transform $\mathbf{T}_{\mathcal{O}\mathcal{L}}$, the LiDAR pose in $\{\mathcal{W}\}$ is:
\begin{equation}
    \mathbf{T}_{\mathcal{W}\mathcal{L}}(t) =
    \mathbf{T}_{\mathcal{W}\mathcal{O}}(t)\,\mathbf{T}_{\mathcal{O}\mathcal{L}}.
\end{equation}
For a 3D point $\mathbf{p}_i\in\mathbb{R}^3$,
\begin{equation}
    \mathbf{z}_i(t) =
    \pi\!\left(\mathbf{T}_{\mathcal{W}\mathcal{L}}(t)^{-1}\,\mathbf{p}_i\right) + \mathbf{n}_i(t),
\end{equation}
where $\pi(\cdot)$ is the projection model and $\mathbf{n}_i(t)$ is measurement noise.

\noindent \textbf{Passive Excitation:}  
Given the above coupling, any acceleration, braking, or turning of the internal drive shifts the system's center of mass in $\{\mathcal{O}\}$, generating gravity and contact torques that perturb $\boldsymbol{\omega}_{\mathcal{O}}(t)$ and hence $\mathbf{R}_{\mathcal{W}\mathcal{O}}(t)$.  
These natural attitude variations in turn alter $\mathbf{T}_{\mathcal{W}\mathcal{L}}(t)$, diversifying scan directions and intermittently exposing near-ground regions without explicit motion shaping.

\noindent \textbf{Control Objective:}  
We aim to generate wheel commands $\mathbf{u}(t)$ so that the induced trajectory $\mathbf{x}(t)$ follows a desired path $\mathbf{x}^\ast(t)$, while state estimation from LiDAR–IMU data leverages the passive excitation to enrich $\{\mathbf{z}_i(t)\}$ and improve odometry robustness.

\subsection{Passive LiDAR Excitation via Spherical Dynamics}
\label{sec:passive_excitation}

Let $\mathbf{u}(t) = [u_L(t),u_R(t)]^\top$ denote the left and right wheel angular velocities of the internal drive.  
From the differential-drive kinematics, the forward (longitudinal) velocity is $\frac{r}{2}\big(u_L(t)+u_R(t)\big)$,  
and the yaw rate (producing lateral acceleration in the shell frame) is $\frac{r}{d}\big(u_R(t)-u_L(t)\big)$.  
These motions of the internal drive shift the system’s instantaneous center of mass in $\{\mathcal{O}\}$,  
resulting in a time-varying offset $\mathbf{c}_{\mathcal{O}}(t)\in\mathbb{R}^3$ that generates a gravity/contact torque:

\begin{equation}
    \boldsymbol{\tau}_g(t)=\mathbf{c}_{\mathcal{O}}(t)\times m\,\mathbf{g}_{\mathcal{O}} .
\end{equation}
Rigid-body attitude dynamics of the shell follow
\begin{equation}
    \mathbf{I}_{\mathcal{O}}\,\dot{\boldsymbol{\omega}}_{\mathcal{O}}(t)
    = \boldsymbol{\tau}_g(t)+\boldsymbol{\tau}_d(t)
      - \boldsymbol{\omega}_{\mathcal{O}}(t)\times\big(\mathbf{I}_{\mathcal{O}}\boldsymbol{\omega}_{\mathcal{O}}(t)\big),
\end{equation}
\begin{equation}
    \dot{\mathbf{R}}_{\mathcal{W}\mathcal{O}}(t)
    = \mathbf{R}_{\mathcal{W}\mathcal{O}}(t)\,\lfloor \boldsymbol{\omega}_{\mathcal{O}}(t)\rfloor ,
\end{equation}
where \(\boldsymbol{\tau}_d(t)\) denotes drive-induced contact torque.
With fixed extrinsics \(\mathbf{T}_{\mathcal{O}\mathcal{L}}=\big[\mathbf{R}_{\mathcal{O}\mathcal{L}},\,\mathbf{p}_{\mathcal{O}\mathcal{L}}\big]\),
the LiDAR pose is
\begin{equation}
\begin{aligned}
    \mathbf{T}_{\mathcal{W}\mathcal{L}}(t) 
    &= \mathbf{T}_{\mathcal{W}\mathcal{O}}(t)\,\mathbf{T}_{\mathcal{O}\mathcal{L}} \\
    &= \begin{bmatrix}
        \mathbf{R}_{\mathcal{W}\mathcal{O}}(t)\mathbf{R}_{\mathcal{O}\mathcal{L}} &
        \mathbf{p}_{\mathcal{W}\mathcal{O}}(t) + 
        \mathbf{R}_{\mathcal{W}\mathcal{O}}(t)\mathbf{p}_{\mathcal{O}\mathcal{L}}\\
        \mathbf{0}^\top & 1
    \end{bmatrix}.
\end{aligned}
\end{equation}
LiDAR measurements for \(\mathbf{p}_i^{\mathcal{W}}\in\mathbb{R}^3\) are modeled as
\begin{equation}
    \mathbf{z}_i(t) = \pi\!\left(\mathbf{T}_{\mathcal{W}\mathcal{L}}(t)^{-1}\mathbf{p}_i^{\mathcal{W}}\right)+\mathbf{n}_i(t).
\end{equation}
Passive excitation thus varies \(\mathbf{R}_{\mathcal{W}\mathcal{O}}(t)\), diversifying scan directions and intermittently exposing near-ground regions without added actuation.

\subsection{Discrete-Time State Estimation}
\label{sec:lio_backend}

At LiDAR frame $k$, the state
\[
\mathbf{x}_k = [\,\mathbf{p}_k,\,\mathbf{v}_k,\,\mathbf{R}_k,\,\mathbf{b}_{a,k},\,\mathbf{b}_{\omega,k}\,]
\]
contains position, velocity, orientation, and IMU biases, which are also used for PID control.  
IMU data are preintegrated \cite{forster2015imu} to form relative motion constraints, while LiDAR scans $\mathcal{Z}_k$ are registered to the map to yield residuals $\mathbf{r}_{L,k}$.  
The backend minimizes
\[
\min_{\{\mathbf{x}_k\}}\ \sum_k \|\mathbf{r}_{L,k}\|^2_{\Sigma_L} + \|\mathbf{r}_{I,k,k+1}\|^2_{\Sigma_I},
\]
where $\mathbf{r}_{I,k,k+1}$ is the IMU preintegration residual.  
Passive rocking enriches $\mathcal{Z}_k$ with vertical and near-ground geometry, improving FAST-LIO alignment without added actuation.

\subsection{Trajectory Control}
Let $\mathbf{x}^E_k$ denote the state estimated by the LiDAR--IMU odometry 
at LiDAR frame $k$ (update rate $\approx 10\ \mathrm{Hz}$), and 
$\mathbf{x}^\ast_n$ the desired state at control cycle $n$ ($100\ \mathrm{Hz}$).  
Between LiDAR updates, the most recent $\mathbf{x}^E_k$ is held constant or interpolated 
to obtain the control state $\mathbf{x}^C_n$.  

The discrete-time tracking error is
\[
    \mathbf{e}_n = \mathbf{x}^\ast_n - \mathbf{x}^C_n ,
\]
which is used in a PID controller running at $100\ \mathrm{Hz}$ to compute 
the target linear and angular velocities for the internal differential drive.  
These velocities are converted into wheel commands
\[
    \mathbf{u}_n = [\,u_L(n),\, u_R(n)\,]^\top ,
\]
enabling the platform to follow the reference trajectory while allowing 
its inherent dynamics to induce passive LiDAR excitation.

\begin{table*}[t]
\centering
\caption{System-level comparison across different robot platforms.}
\label{tab:system_comparison}
\begin{tabular}{l|c|c|c|c|c|c}
\hline
\hline
\textbf{Platform} & \textbf{Map Completeness} & \textbf{Mean Tracking Error} & \textbf{Power} & \textbf{Cost} & \textbf{Weight} & \textbf{Size} \\
\hline
\makecell[l]{Mecanum Wheel Robot\\ Fixed Horizontal LiDAR} 
 & \makecell{Low \\ 63\%} 
 & \makecell{High \\ 0.137} 
 & \makecell{Low \\ 32 W} 
 & \makecell{Low \\ 2.3k USD} 
 & \makecell{Medium \\ 3.8 kg} 
 & \makecell{Small \\ $26\times23\times16$ cm} \\
\hline
% \makecell[l]{Mecanum Wheel Robot\\ Fixed tilted LiDAR} 
%  & \makecell{Low \\ 86\%} 
%  & \makecell{High \\ ($\sim$0.13--0.15 m)} 
%  & \makecell{Low \\ ($\sim$20 W)} 
%  & \makecell{Low \\ ($<$2k USD)} 
%  & \makecell{Medium \\ ($\sim$15 kg)} 
%  & \makecell{Small \\ ($40\times40\times30$ cm)} \\
% \hline
\makecell[l]{Quadruped Robot \\ Self-Rotating LiDAR} 
 & \makecell{Full Coverage \\ \textbf{100}\%} 
 & \makecell{Highest \\ 0.172m} 
 & \makecell{Very high \\ 170 W} 
 & \makecell{Very high \\ 13k USD} 
 & \makecell{High \\ 15 kg} 
 & \makecell{Large \\ $65\times28\times78$ cm} \\
\hline
\makecell[l]{Proposed (PERAL) \\ Self-Excitation LiDAR} 
 & \makecell{High \\ 96\%} 
 & \makecell{Lowest \\ \textbf{0.10 m}} 
 & \makecell{Low \\ \textbf{22} W} 
 & \makecell{Low \\\textbf{ 2.1k} USD} 
 & \makecell{Low \\ \textbf{1.8 }kg} 
 & \makecell{Compact \\ $25\times25\times25$ cm} \\
\hline
\end{tabular}
\end{table*}

\section{Experiments}
~\label{sec:exp_setup}

\subsection{Experiments Objectives}
The experiments aim to validate three key aspects of the proposed \emph{PERAL} spherical robot:
\begin{enumerate}
    \item \textbf{Perceptual effectiveness:} quantify the improvement in LiDAR coverage and map completeness compared to conventional platforms.
    \item \textbf{Controllability:} verify that passive excitation does not compromise motion stability, as evidenced by trajectory tracking accuracy.
    \item \textbf{System-level efficiency:} evaluate trade-offs in power, weight, and cost against active-rotation and fixed-LiDAR baselines.
\end{enumerate}
In addition, we demonstrate the potential of the \emph{PERAL} robot for search-and-rescue missions in civil defense scenarios.

\subsection{Experimental Setup}
The experimental platform is a custom-built spherical mobile robot with an internal differential drive unit, equipped with a DJI Livox Mid-360 LiDAR and an onboard IMU. 
A LubanCat RK3588 single-board computer running ROS~2 handles data acquisition and processing. 
The tests were conducted in three representative environments:
\begin{itemize}
    \item \textbf{Indoor laboratory:} cluttered space with desks, chairs, and equipment.
    \item \textbf{Corridor:} long, narrow passages with sparse geometric features.
    \item \textbf{Tactical training center:} environment with obstacles and a sloped floor to evaluate operation on inclined terrain.
\end{itemize}

Conducting hardware-level comparisons is particularly challenging, as they must balance fairness, feasibility, and performance tuning. Within the scope of our available platforms, we benchmark three representative configurations:
\begin{itemize}
    \item {Mecanum Wheel Robot with Fixed Horizontal LiDAR \cite{shafiq2024real}}.
    \item {Quadruped Robot with Self-Rotating LiDAR\cite{Li2025UAMPC}}.
    \item {Proposed PERAL Spherical Robot with Self-Excitation LiDAR}.
\end{itemize}
All platforms employ the same LiDAR–IMU sensor unit and run FAST-LIO2 for odometry and mapping, ensuring fair comparison.

\subsection{Evaluation Metrics}
Evaluation combines quantitative and qualitative measures:
\begin{itemize}
    \item \textbf{Trajectory tracking error:} mean deviation from a predefined reference trajectory (figure-8, O, oval).
    \item \textbf{Map completeness:} voxel-level recall ratio $C$ against Leica terrestrial laser scans.
    \item \textbf{Near-ground visibility:} visual inspection of additional ground returns not seen by strictly horizontal LiDAR.
    \item \textbf{Qualitative robustness:} observation of SLAM trajectory continuity and loop closure success in feature-sparse settings such as corridors.
    \item \textbf{System efficiency:} comparison of power, cost, weight, and volume across platforms (Tab.~\ref{tab:system_comparison}).
\end{itemize}

\begin{figure*}[t]
    \centering
    \includegraphics[width=1\linewidth]{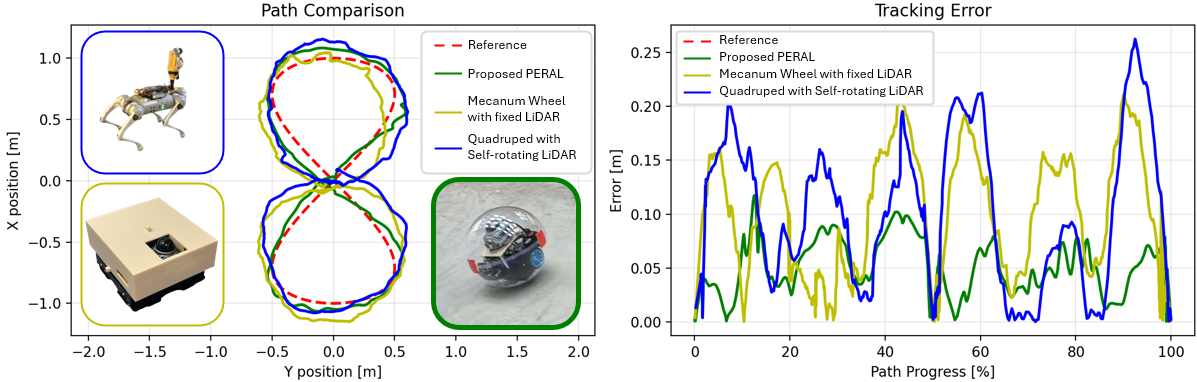}
    \caption{Path tracking performance in a figure-8 trajectory. 
(Left) Comparison between the reference trajectory and three robot platforms: the proposed PERAL spherical robot, an omnidirectional robot, and a quadruped robot. 
(Right) Tracking error along the trajectory progress, showing that the proposed robot achieves lower average error compared to the baselines.}
    \label{fig:trajectorytrackingresult}
\end{figure*}

\subsection{Experimental Procedure}
Each platform is evaluated under identical conditions to ensure fairness. 
Robots are commanded to follow predefined reference paths at comparable speeds ($0.4$--$0.6~\mathrm{m/s}$). 
During each run, LiDAR and IMU measurements are collected and processed through the same FAST-LIO2 pipeline for odometry and mapping. 
For statistical reliability, every configuration is repeated three times. 
Results are then analyzed quantitatively (trajectory tracking error and voxel-based completeness ratio $C$) and qualitatively (point cloud density in near-ground regions, robustness of SLAM trajectories).

\subsection{Trajectory Tracking Tests}
Controllability is assessed by executing three types of reference trajectories:
\begin{itemize}
    \item \textbf{Figure-8:} two tangent circles of diameter $1~\mathrm{m}$, producing alternating straight and curved motion within a compact $1 \times 2~\mathrm{m}$ footprint.
    \item \textbf{Figure-Circle:} a single circle ($R=1~\mathrm{m}$), evaluating steady turning and uniform excitation.
    \item \textbf{Figure-Oval:} an ellipse ($2 \times 4~\mathrm{m}$), combining long straights with gentle curves to test mixed motion modes.
\end{itemize}
The deviation between estimated and reference paths is computed as mean absolute tracking error. 
As shown in Fig.~\ref{fig:trajectorytrackingresult}, the \textbf{Proposed PERAL Spherical Robot (Self-Excitation LiDAR)} achieves the lowest error ($\sim$0.10 m), compared with the \textbf{Mecanum Wheel Robot (Fixed Horizontal LiDAR)} ($\sim$0.137 m) and the \textbf{Quadruped Robot (Self-Rotating LiDAR)} ($\sim$0.172 m). 
These results confirm that passive excitation does not compromise motion stability and can even enhance tracking robustness under curved segments.

\subsection{Map Completeness Tests}
\begin{figure}[t]
    \centering
    \includegraphics[width=0.95\linewidth]{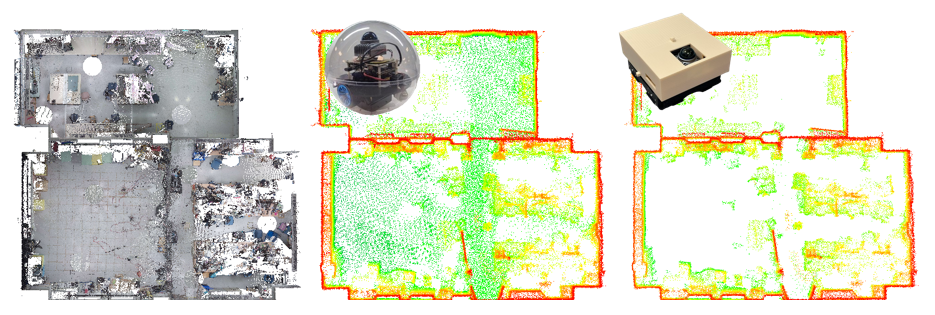}
\caption{Map reconstruction comparison across platforms. 
Left: Leica ground-truth scan. 
Middle: proposed PERAL spherical robot ($\sim$96\% completeness). 
Right: omnidirectional robot with fixed horizontal LiDAR ($\sim$63\%). 
For reference, a quadruped with an actively rotating LiDAR can achieve 
100\% coverage, but only at significantly higher cost and complexity.}
    \label{fig:mapcoverage}
\end{figure}

Map completeness $C$ is computed as
\[
C = \frac{|\mathcal{V}_{\text{ref}} \cap \mathcal{V}_{\text{est}}|}{|\mathcal{V}_{\text{ref}}|},
\]
where $\mathcal{V}_{\text{ref}}$ and $\mathcal{V}_{\text{est}}$ denote occupied voxels in the Leica reference and reconstructed maps, respectively.  
Fig.~\ref{fig:mapcoverage} shows that PERAL achieves $\sim$96\% completeness, significantly higher than the fixed horizontal baseline ($\sim$63\%) and close to the active rotating LiDAR ($100\%$). 
Qualitative inspection further confirms that PERAL recovers near-ground and vertical structures without requiring mechanical actuation.

\begin{figure}[t]
    \centering
    \includegraphics[width=0.95\linewidth]{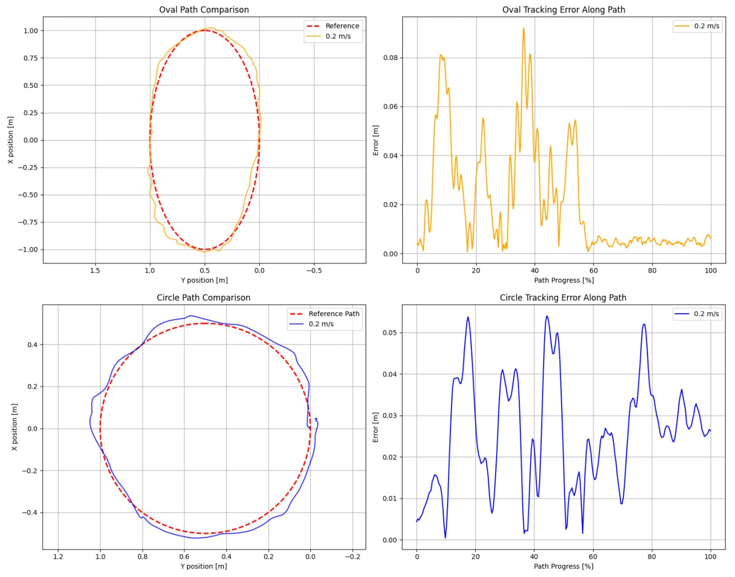}
\caption{Trajectory tracking tests of the PERAL for running a figure circle and figure oval.}
    \label{fig:trajectorytrackingtestmore}
\end{figure}

\section{Results and Discussion}

\subsection{Trajectory Tracking Performance}
As shown in Fig.~\ref{fig:trajectorytrackingresult} and Fig.    \ref{fig:trajectorytrackingtestmore}, the 
\textbf{Proposed PERAL Spherical Robot (Self-Excitation LiDAR)} 
achieves the lowest mean tracking error ($\sim$0.10 m), compared with the 
\textbf{Mecanum Wheel Robot (Fixed Horizontal LiDAR)} ($\sim$0.137 m) 
and the \textbf{Quadruped Robot (Self-Rotating LiDAR)} ($\sim$0.172 m). For certain trajectories, PERAL achieves even lower error.
This indicates that passive excitation preserves controllability and can 
even enhance stability under curved segments, avoiding error peaks 
observed in the baselines.

\subsection{Map Completeness}
Map completeness
shows clear differences across platforms. 
The \textbf{Mecanum Wheel Robot with fixed horizontal LiDAR} achieves only 
$\sim$63\%, as near-ground regions remain largely unobserved. 
The \textbf{Quadruped Robot with self-rotating LiDAR} reaches full coverage 
(100\%), but this requires additional actuation and comes at the cost of 
very high power, weight, and expense. 
In contrast, the \textbf{Proposed PERAL Spherical Robot (Self-Excitation LiDAR)} 
achieves $\sim$96\% completeness, successfully capturing low-height structures 
and wall continuity without added mechanical complexity. 
This result demonstrates that passive excitation provides coverage close to 
active scanning while maintaining efficiency and compactness.

\subsection{System-Level Comparison}
Tab.~\ref{tab:system_comparison} highlights the trade-offs between 
coverage and efficiency. The Mecanum baseline offers low cost and 
power but limited completeness. The Quadruped achieves full coverage 
but requires $>$100 W, weighs 15 kg, and costs over 10k USD. 
The PERAL robot balances both ends: near-full completeness ($\sim$96\%), 
lowest tracking error, and compact size ($1.8$ kg, $25 \times 25 \times 25$ cm) 
with modest power (22 W). This demonstrates that passive excitation 
provides an effective compromise between fidelity and efficiency.

\subsection{Slope Crossing Capability}
\begin{figure}[h]
    \centering
    \includegraphics[width=0.95\linewidth]{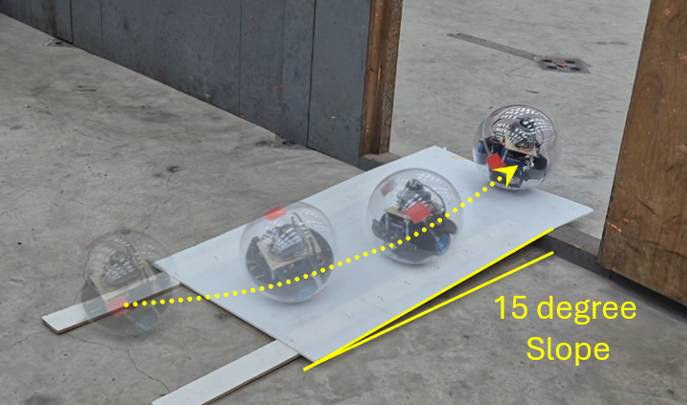}
\caption{PERAL spherical robot climbing a sloped ramp ($\sim$15°), demonstrating stable motion on inclined terrain in the tactical training center.}
    \label{fig:slopecrossing}
\end{figure}

To further validate mobility, we tested the PERAL robot on an inclined ramp inside the tactical training center (Fig.~\ref{fig:slopecrossing}). 
The robot successfully ascended a slope of approximately $14^{\circ}$ without loss of stability or slippage. 
This demonstrates that the passive self-excitation mechanism does not hinder basic locomotion performance, and the spherical design preserves sufficient ground traction to handle moderate inclines. 
Such capability is important for mixed indoor--outdoor or semi-structured environments, where uneven terrain or ramps are common.

\subsection{Ground-Level Target Detection and Applications}
\begin{figure}[h]
    \centering
    \includegraphics[width=0.95\linewidth]{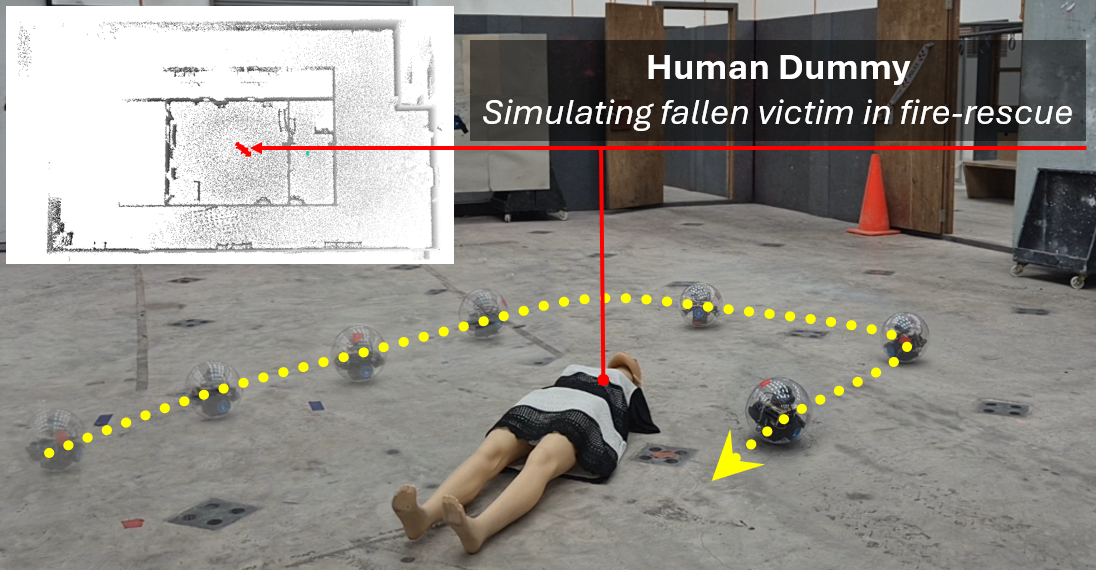}
\caption{Human detection experiment in the tactical training center. 
The proposed \emph{PERAL} spherical robot successfully identifies a human dummy lying on the ground, as highlighted in both the SLAM map (inset) and the reconstructed trajectory (yellow path). This demonstrates that passive LiDAR excitation enables the capture of near-ground structures, making otherwise occluded or low-height obstacles detectable.}
    \label{fig:humandummy}
\end{figure}

A critical advantage of improved near-ground coverage is the ability to detect and localize targets lying on the floor, which are often missed by conventional horizontal LiDAR configurations. 
To demonstrate this capability, we placed a human-sized dummy in the tactical training center and commanded the robot to navigate a loop trajectory around the target (see Fig.~\ref{fig:humandummy}).

As the proposed \emph{PERAL} spherical robot performs passive self-excitation, its LiDAR captures additional low-height returns corresponding to the dummy. 
These points are integrated into the SLAM map, allowing the system to localize the dummy's presence without requiring any dedicated perception module. 
The inset of Fig.~\ref{fig:humandummy} shows the reconstructed map, where the dummy appears as a cluster of occupied voxels, confirming that near-ground observability directly enables ground-level human detection. 

This capability is particularly relevant to \emph{search-and-rescue} and \emph{firefighting} applications, where victims may collapse onto the ground and remain outside the field of view of traditional LiDAR placements. 
By leveraging passive excitation to capture these low-height structures, PERAL expands the perceptual horizon of compact robots, enabling them to contribute to early victim detection and situational awareness in hazardous environments.

\subsection{Discussion}
Overall, results confirm that passive LiDAR excitation enriches vertical 
scan diversity, improving near-ground visibility and map completeness 
while preserving trajectory tracking accuracy. The approach achieves 
coverage close to active rotation, but with much lower cost, weight, and 
power. This makes PERAL particularly suitable for compact, 
energy-constrained robots in structured environments (labs, corridors) 
and tactical domains where slope and uneven ground must be captured.

\section{Conclusion}
This paper presents a passive LiDAR excitation strategy for spherical robots that exploits natural rocking motions during locomotion to intermittently reorient the scanning plane, improving near-ground coverage without additional actuators. Experiments in laboratory, corridor, and tactical settings show reduced blind zones and enriched low-height map details, achieving coverage comparable to active rotation while preserving simplicity and efficiency. Future work will focus on formal observability analysis and on reinforcement learning and data-driven optimization \cite{messikommer2024reinforcement} to tune motion parameters or generate excitation-rich trajectories for enhanced robustness across diverse environments.

\section*{Acknowledgments}
This work was developed with assistance from OpenAI's GPT-4. 
The system was used for (i) generating and refining code for the integration of robot functions, including autonomous control modules, 
and (ii) drafting and editing sections of the manuscript. 
All AI-generated content was reviewed and verified by the authors.

%\addtolength{\textheight}{-12cm}   % This command serves to balance the column lengths
                                  % on the last page of the document manually. It shortens
                                  % the textheight of the last page by a suitable amount.
                                  % This command does not take effect until the next page
                                  % so it should come on the page before the last. Make
                                  % sure that you do not shorten the textheight too much.

%%%%%%%%%%%%%%%%%%%%%%%%%%%%%%%%%%%%%%%%%%%%%%%%%%%%%%%%%%%%%%%%%%%%%%%%%%%%%%%%

%%%%%%%%%%%%%%%%%%%%%%%%%%%%%%%%%%%%%%%%%%%%%%%%%%%%%%%%%%%%%%%%%%%%%%%%%%%%%%%%

%%%%%%%%%%%%%%%%%%%%%%%%%%%%%%%%%%%%%%%%%%%%%%%%%%%%%%%%%%%%%%%%%%%%%%%%%%%%%%%%

\tiny
\bibliographystyle{IEEEtran}
\bibliography{refs}

\end{document}